\documentclass[runningheads]{llncs}

\usepackage[T1]{fontenc}
\usepackage[utf8]{inputenc}
\usepackage[russian,main=english]{babel}

\usepackage{amsmath,amssymb,amsfonts,amsbsy,amstext,amscd,amsxtra}

\usepackage{url}
\usepackage{graphicx}
\usepackage{xcolor}

\newcommand*{\hm}[1]{#1\nobreak\discretionary{}{\hbox{$\mathsurround=0pt #1$}}{}}

\usepackage{mathtools}

\DeclarePairedDelimiterX{\infdivx}[2]{(}{)}{%
  #1\;\delimsize\|\;#2%
}
\newcommand{\infdiv}{\infdivx}

\usepackage{booktabs}
\usepackage{csquotes}

\usepackage[detect-all]{siunitx}
\begin{document}

\title{Determination of the Number of Topics Intrinsically: Is It Possible?}

\author{%
  Victor Bulatov\inst{1}\orcidID{0000-0002-2026-9774} \and
  Vasiliy Alekseev\inst{1}\orcidID{0000-0001-7930-3650} \and
  Konstantin Vorontsov\inst{2}\orcidID{0000-0002-4244-4270}
}

\institute{%
    Moscow Institute of Physics and Technology, Dolgoprudny, Russia\\
    \email{\{viktor.bulatov,vasiliy.alekseyev\}@phystech.edu} \and
    Lomonosov Moscow State University, Moscow, Russia\\
    \email{vokov@forecsys.ru}
}

\maketitle

\begin{abstract}
The number of topics might be the most important parameter of a topic model.
The topic modelling community has developed a set of various procedures to estimate the number of topics in a dataset, but there has not yet been a sufficiently complete comparison of existing practices.
This study attempts to partially fill this gap by investigating the performance of various methods applied to several topic models on a number of publicly available corpora.
Further analysis demonstrates that intrinsic methods are far from being reliable and accurate tools.
The number of topics is shown to be a method- and a model-dependent quantity, as opposed to being an absolute property of a particular corpus.
We conclude that other methods for dealing with this problem should be developed and suggest some promising directions for further research.
 
\keywords{Topic modelling \and Number of topics \and Coherence \and Diversity \and Perplexity \and Stability \and Entropy}

\end{abstract}

\section{Introduction}

Topic models are statistical models which are usually employed for unsupervised text analysis.
Topic modelling assumes that there are a number of latent topics which explain the collection.
Following the convention, we will denote the number of documents by $D$, the number of topics by $T$ and the size of vocabulary by $W$.

The topic model is trained by inferring two probability distributions: the ``word-in-topic`` distribution (colloquially referred to as $\phi_{wt}: = p(w \mid t)$ or as an column of a stochastic matrix $\Phi$ with the shape $W \times T$) and the ``topic-in-document`` distribution (colloquially referred to as $\theta_{td} := p(t \mid d)$ or as a row of a stochastic matrix $\Theta$ with the shape $T \times D$).
We limit our discussion to $\Phi,~\Theta$ parameter matrices since they are present in every topic model.
Some topic models introduce additional parameters beside these; exploring intrinsic measures related to these parameters is beyond the scope of this work.

The number of topics $T$ is the key hyperparameter of the most topic models.
Naturally, there are a number of influential publications suggesting a way to select this hyperparameter  \cite{griffiths2004finding,Brunet4164,cao2009density,arun2010finding,deveaud2014accurate,greene14howmany,zhao15heuristic}.
However, there is no accepted consensus on this matter.
Namely, there is no agreement in the literature on the sequence of steps one must carry out to determine the best number of topics.
Also, there appears to be a disagreement on which methods are appropriate \cite{soleimani14parsimonious,krasnov19clustering}.

We are most interested in \textit{intrinsic} metrics which do not use any external resources, labels or human assessment.
This approach is typically based on presenting different models with different numbers of topics, obtaining a mea\-sure\-ment of a certain quality metric (possibly using cross-validation on held-out document sets) and selecting the number of topics corresponding to the best value. We do not explore extrinsic approaches aimed at optimizing some external criterion of interest, such as classification using a labelled validation dataset, because it is clear that explicitly optimizing for some secondary task provides a better result as measured by the model performance on that task.


Notably, we exclude models which learn the required number of topics au\-to\-mat\-i\-cal\-ly such as hierarchical Dirichlet process \cite{teh2005sharing,bryant2012truly}.
We elected to exclude them due to the two reasons.
First, they tend to add a new set of hyperparameters that require optimization.
Second, they are not universal: any topic model containing $\Phi$ and $\Theta$ distributions could be scored according to any internal metric, while complex Bayesian topic models optimize some loss function specific to their parameters\footnote{That being said, such models sometimes contain model-agnostic metrics which they are implicitly optimizing. We will use these quality metrics if possible.}.


This work investigates the number of metrics proposed in literature over the range of different corpora and different topic models and attempts to formulate a set of useful guidelines for practitioners.

The paper is structured as follows. In Section \ref{sec:related-work}, we review papers related to our methodology on the whole, while Section \ref{sec:quality-metrics} reviews and assesses various methods proposed to choose the ``right'’ number of topics.
Section \ref{sec:methodology} describes the design of the experiment; namely, topic models and corpora used.
In section \ref{sec:methodology}, we give experimental evidence related to the issue of $T$ determination. Finally, we conclude in section \ref{sec:results-and-discussion} with a few points of discussion.


\section{Related Work}
\label{sec:related-work}

Many researchers proposed a way to determine $T$ (see section \ref{sec:quality-metrics}).
Often, they perform a very limited survey of other existing approaches.
Also, it should be noted that their experiments are often limited to certain topic model families or by the range and/or size of datasets involved.
In this section, we are discussing previous work that reviews several approaches.

Known to us, two software packages implement a number of metrics and operate on a similar idea:  they allow the user to explore the values of several quality metrics among the range of different $T$ and select a point that appears most advantageous.

The \texttt{ldatuning} package\footnote{https://github.com/nikita-moor/ldatuning} for R \cite{ldatuning} has 4 methods supported: D-Spectral \cite{arun2010finding}, D-avg-COS \cite{cao2009density}, D-avg-JS \cite{deveaud2014accurate}, and holdPerp \cite{griffiths2004finding}.
It depends on the \texttt{topicmodels} package, which supports LDA and CTM models.
The recent work of \cite{hou2018benchmarking} examines the performance of these methods on several generated datasets with a known value of $T$.

The TOM library\footnote{https://github.com/AdrienGuille/TOM} for Python \cite{guille2016tom} implements 3 methods for estimation of $T$: D-Spectral \cite{arun2010finding}, toptokens-ssample-stab \cite{greene14howmany}, and cophenet \cite{Brunet4164}.
It supports Latent Dirichlet Allocation models and models based on Non-negative Matrix Fac\-tor\-iza\-tion.

Another comparable work is \cite{gialampoukidis2016hybrid} that proposes a new process for estimating the number of clusters in a dataset and compares it to a large number of traditional metrics as implemented in \texttt{NbClust} R package.
The comparison is evaluated on 20NG corpus and several subsets of WikiRef220 dataset; most of traditional metrics fail to obtain good results.

The work of \cite{krasnov19clustering} examines and enumerates a number of methods.


\section{Intrinsic Quality Metrics}
\label{sec:quality-metrics}

This paper mainly focuses on intrinsic metrics, which will be discussed in this section.
The section is organized into several broad categories, each category containing a number of somewhat related ideas found in the literature. 

\subsection{Perplexity}
The classic intrinsic approach is based on hold-out perplexity \cite{griffiths2004finding} (\textbf{holdPerp}).
The work of \cite{zhao15heuristic} enhances this method by considering rates of perplexity change (\textbf{RPC}) instead of raw perplexities (essentially, they take into account the slope of perplexity curve instead of absolute values).

\subsection{Stability}
In \cite{greene14howmany,belford2018stability}, authors employ stability analysis to judge the quality of modelling choices (most notably, the number of topics).
This approach is often used when analysing clustering models such as k-means or Non-negative Matrix Fac\-tor\-iza\-tion (NMF).
Intuitively, solutions with the ``incorrect`` number of clusters are unstable since they are forced to either merge clusters in an arbitrary way or to create arbitrary partitions of data. 

The stability (\textbf{toptokens-ssample-stab}) is measured by repeatedly creating a shuffled subsample of data, fitting a topic model to it, and then comparing the top-tokens of models created that way to the top-tokens of reference model build on the entire dataset.
The numeric value is calculated as the Jaccard Similarity Index of the assignment obtained by the Hungarian algorithm.

Another important work among the same lines is \cite{Brunet4164} where authors focus on the reproducibility of the class assignments.
Each document $d$ is assigned to its most probable topic; then for each pair $(d_1, d_2)$ it is recorded if $d_1$ and $d_2$ belong to the same cluster; the result is $D \times D$ connectivity matrix.
This matrix is averaged among a number of different topic models.
The proposed measure of stability is defined as a cophenetic correlation coefficient of the average connectivity matrix. 

Note that subsampling was not employed; the randomization comes from different initializations only.
The TOM library uses ten random runs by default.
Unfortunately, we could not test the validity of this approach, due to the heavy computational cost demanded by training additional topic models on large corpora involved.


\subsection{Diversity and Sufficiency}
The idea here is that the number of topics should enable the model to describe corpus adequately, but when the number of topics is too large, the model produces a lot of small topics similar to each other.

The approach proposed in \cite{plavin} starts with an excessively large number of topics and then uses regularization to set most of them to zero.
Notably, the proposed regularizer is able to remove linear combinations of existing topics from the model.
The function being optimized by the regularizer is:

\[
\textup{KL}\infdiv*{u(t)}{p(t)}=\textup{KL}\infdiv*{\frac{1}{T}}{\sum_d \theta_{td} \frac{n_d}{n}} \rightarrow \max,
\]

where $t$ refers to a topic inferred by the model.
This is Kullback--Leibler divergence between uniform $u(t) = \frac{1}{T}$ distribution and $p(t)$ as inferred with the help of $\Theta$ matrix.
This could be used as a quality criterion \textbf{UniThetaDivergence}.

\cite{tang14look} takes this process in the opposite direction.
They start with a small number of topics and iteratively add new topics describing documents which are poorly explained by the model.
This approach trades the unknown $T$ for unknown threshold $\epsilon$.
Authors suggest using overall topics diversity as a criterion for determining $\epsilon$.

The use of diversity is based on the intuition that the number of topics is connected to the granularity of topics: when the number of topics is too large, the model produces a lot of small topics similar to each other.
The most influential work along these lines \cite{cao2009density} proposes the usage of average cosine distance between topics (\textbf{D-avg-COS}) as a criterion for model selection.
This idea is expanded in \cite{deveaud2014accurate} by considering Jensen-Shannon divergence (\textbf{D-avg-JS}) instead of cosine distance.
The work of \cite{tang14look}, mentioned above, employs another variant of diversity, average Euclidean distance (\textbf{D-avg-L2}).

In this work, we expand on existing methodology by considering the average distance to the closest topic (instead of average pairwise distance); as a result, \textbf{D-cls-COS}, \textbf{D-cls-JS}, and \textbf{D-cls-L2} are obtained.
In addition, we employ \textbf{D-avg-H} and \textbf{D-cls-H}, which are based on Hellinger distance.

Another important development (\textbf{D-Spectral}) was proposed in \cite{arun2010finding} which integrates information in $\Phi$ and $\Theta$ matrices by considering spectral values of $\Phi$ and rows of unnormalized $\Theta$.
The proposed \textit{Spectral Divergence Measure} reflects the degree of orthogonality between topic vectors.

\subsection{Clustering}

One can also employ a number of metrics usually associated with network analysis and clustering analysis, notably Silhouette Coefficient (\textbf{SilhC}) and Calinski-Harabasz Index (\textbf{CHI}) \cite{mehta_clustering_bank,panichella2013effectively,karami2018fuzzy}.
Interestingly, Krasnov et al. \cite{krasnov19clustering} fail to reproduce Silhouette and Calinski--Harabasz results on a particular dataset.

\subsection{Information-Theoretic Criteria}

Another method is the usage of Bayesian Information Criterion (\textbf{BIC}) which balances the goodness of fit and model complexity.
The most notable recent work making use of this is \cite{soleimani14parsimonious} where BIC is greatly expanded on, and a form of BIC is derived that accounts for additional parameters of a new proposed model.
Minimum Description Length (\textbf{MDL}) formalism was also used \cite{image_segm,gerlach2018network} as well as MML \cite{mml}.
The work of \cite{than2012fully} explores Akaike Information Criterion (\textbf{AIC}) and BIC as a function of the number of topics, although it is not the main focus of this paper.

To calculate these metrics, we proceed as follows.
First, we obtain $\mathfrak{L}(\Phi,\Theta)$, a model likelihood.
Second, we need to find out the number of free parameters $N_p$, which could be calculated in two different ways: the dimensions of $\Phi$ or the number of non-zero entries of $\Phi$ which we denote by $\#\Phi$ (note \cite{than2012fully} argue that a number of free parameters in LDA and sparse models should be treated differently).
The following expressions~(\ref{tab:aic-bic-mdl-calc}) summarize our approach.

\begin{table}[h]
    \centering
    \caption{Equations for calculating AIC, BIC, MDL metrics.}
    \label{tab:aic-bic-mdl-calc}
    \begin{tabular}{lll}
    \toprule
               & $N_p$       & Formula                                             \\
    \midrule
    \textbf{sparse AIC} & $\#\Phi$       & $2 N_p - 2 \mathfrak{L}$                \\
    \textbf{AIC}        & $(W - 1) * T$  & $2 N_p - 2 \mathfrak{L}$                \\
    \textbf{sparse BIC} & $\#\Phi$       & $N_p \log(D) - 2 \mathfrak{L}$          \\
    \textbf{BIC}        & $(W - 1) * T$  & $N_p \log(D) - 2 \mathfrak{L}$          \\
    \textbf{sparse MDL} & $\#\Phi$       & $N_p \log(TD) - 2 \mathfrak{L}$         \\
    \textbf{MDL}        & $(W - 1) * T$  & $N_p \log(TD) - 2 \mathfrak{L}$\\
    \bottomrule
    \end{tabular}
\end{table}

\subsection{Entropy}

The work of \cite{koltcov2018application} develops an analogy between topic models and non-equilibrium complex systems where the number of topics is equivalent to the number of states each particle (word) can occupy.
It is suggested that the ``correct'' number of topics should correspond to the equilibrium state, which is characterised by the minimum of entropy. That way the problem is reduced to finding the minimum of a particular function.

To compute entropy, one needs to determine the set $S \hm= \{(w, t) \hm\mid \phi_{wt} \hm> \varepsilon_0 \} \hm\subset \Phi$ for some fixed $\varepsilon_0$.
Afterwards, the energy is defined as $E \hm= -\log \sum_{(w,t) \in S} \phi_{wt}$, free energy as  $E_f \hm= E - T \log\bigl(|S| / (W * T)\bigr)$. Finally, Renyi entropy is calculated as $-E_f / (T - 1)$. 

The work of \cite{koltcov2018application} uses $\varepsilon_0 = (W)^{-1}$, but we found that it did not perform well in some cases.
We elected to consider the cases of $\varepsilon_0 = 2 (W)^{-1}$ and $\varepsilon_0 = 0.5 (W)^{-1}$ as well. All these criteria are denoted by \textbf{renyi-0.5}, \textbf{renyi-1}, and \textbf{renyi-2}.

\subsection{Lift}

This quality measure (\textbf{lift-score}) was introduced in recent work \cite{fan2019assessing}, where it was observed that LDA models with more ``advanced'' informative priors correspond to higher lift-scores.
Hence, lift-score could be helpful for tuning model hyperparameters.
This poses an interesting question of whether lift-score could be used for $T$ determination.

\subsection{Top-Tokens Analysis}

Although not reflected in scientific literature, another reasonable approach is to build many models with different $T$ and pick the one that gives the highest coherence value \cite{mlplus,sathi2016quality,del2020emerging}.
\textit{Coherence} is widely used quality metric for topic models, which is computed by using co-occurrence counts of top $10$ most probable words of each topic (top-tokens).

Krasnov et al.\cite{krasnov19clustering} offers an interesting variation: top $10$ words were replaced with their dense embeddings (GloVe representations were used) and the number of topics was successfully chosen in accordance with Davies--Bouldin index, an indicator of clustering quality.

\section{Methodology}
\label{sec:methodology}

The question we aim to answer: is the notion of ``optimal number of topics`` well-defined?
In other words, is there an agreement among different approaches proposed in literature?
Also, are proposed approaches sufficiently robust to the parameters change?

Our methodology is as follows.
We train a number of different topic models (such as PLSA, LDA, ARTM, and TARTM \ref{sec:topicmodelstudied}) with the $T$ hyperparameter ranging from $T_{min}$ to $T_{max}$ for a number of iterations sufficient for convergence.
The training process starts from three different random initializations, and several quality metrics are measured for each run.
If a method requires a value of parameter to be chosen by user, we will then explore various candidate values. 

We repeat this process for many different corpora. To allow the calculation of held-out perplexity, each corpus is randomly split into train and test ($80\%$ train, $20\%$ test).
We do not shuffle documents, since the inference algorithm we use (BigARTM\cite{frei2016parallel}) does not depend on the document order.

As a separate experiment, we use stability analysis for determination of $T$. 
Our implementation\footnote{\url{https://github.com/machine-intelligence-laboratory/OptimalNumberOfTopics}}
is adapted from TOM library \cite{guille2016tom}, which is inspired by \cite{greene14howmany,agrawal2018wrong}.
The procedure revolves about assessing topic diversity over subsampled datasets.
Unlike the paper we do not train initial reference model $S_{0}$ and work with all the models obtained for the fixed number of topics.
We create $5$ subsamples of the original dataset without replacement.
The size of all dataset subsamples $D_i \hm\subseteq D$, $i \hm\in [0, S)$ is fixed and set equal to $0.5$ of the size of the original dataset.
We train topic models for a range of topic numbers $t \hm\in [t_{\min}, t_{\max}]$ on each of the obtained subsamples.
A seed which determines initial weights in $\Phi_t$ matrix is fixed and set equal to $0$ for all topic models.
After data subsampling and model training, the last step is comparison of the models with the same numbers of topics $t$ but trained on the different data subsamples $D_i, D_j$, $i \hm{\not=} j$.
For each number of topics $t$, we have $S$ topic models which means that there are $\binom{S}{2}$ possible pairs of topic models in total ($\binom{5}{2} \hm= 10$ in the experiments).
To get a value of distance between topic models, we compare their topics each-to-each using the Jaccard distance function.
Then, having a $|T| \hm\times |T|$ matrix of topic distances and by getting a solution of this linear sum assignment problem, we compute the distance between topic models $\rho_{stab}\bigl(\Phi_t(D_i), \Phi_t(D_j)\bigr)$, $i, j \hm\in [0, S)$.
These distances are then averaged over the number of comparisons of topic models ($10$ comparisons in the experiments).
Thus, we come to the formula which we use to get an instability estimate of the topic model for the number of topics $t \hm\in [t_{\min}, t_{\max}]$:
\[
  \frac{1}{\binom{S}{2}} \sum_{0 \leq i < j < S} \rho_{stab}\bigl(\Phi_t(D_i), \Phi_t(D_j)\bigr)
\]
The lower this number, the better.
This logic differs from \cite{greene14howmany}, where the proposed stability score, on the other hand, is an estimate of the similarity of the models and therefore the higher the better.
Another thing which differs is that our estimate is computed using each-to-each pairwise comparisons of the models trained on different data subsamples.
Whereas \cite{greene14howmany} train a model on a whole dataset first and then use it as a reference point during comparisons.
The models trained on data subsamples are not compared to each other, but to this reference point model.
The obvious advantage of the approach of \cite{greene14howmany} is that it is faster: the number of comparisons is limited by the number of models $S$, while in our case the bound is $\binom{S}{2}$.
On the other hand, the authors in \cite{greene14howmany} do not guarantee that the reference point model is good, which we believe is important.
As for the speed of the computation process, in case of large $S$ the number of model-to-model in our formula can be manually limited to a reasonable number.


Ideally, the value ``recommended'' by some particular method should cor\-re\-spond to the pronounced minimum/maximum on the plot.
This fact motivates our further analysis where we attempt to locate and classify global optima al\-go\-rith\-mi\-cal\-ly as follows.

First, we take note of the highest and lowest points ($h$ and $l$).
We select all points which values fall into $[h - \alpha (h - l), h]$ interval (or $[l, l + \alpha (h - l)]$ if the score should be minimized).
We hold $\alpha=0.07$ in our analysis.
Second, we test whether those points are adjacent to each other (if they are, the optimum is single and robust; otherwise the curve is either jumping or has several significant local optima). Additionally, we check if the optimum was achieved on the boundary of the explored range.


\subsection{Topic Models Studied}
\label{sec:topicmodelstudied}

\textbf{PLSA}.
PLSA \cite{hofmann1999probabilistic} is a simple topic model without any additional hyperparameters aside from $T$.

\textbf{LDA}.
LDA is a well-known topic model, having prior $\eta$ for $\Phi$ distribution and a prior $\alpha$ for $\Theta$ distribution (priors could be numbers or vectors).
We implemented three variants of LDA model inside BigARTM/TopicNet technology stack: \textit{double-symmetric} ($\eta = \alpha = \frac{1}{T}$), \textit{asymmetric} (following the recommendation from \cite{wallach2009rethinking} we use symmetric prior over $\Phi$ and asymmetric over $\Theta$: 
$\eta=\frac{1}{T}$, $\alpha_{td}=\frac{1}{\sqrt{t + T}},~0\leq t \leq T$) and \textit{heuristic} (the values 
$\alpha=\frac{50}{T}$ and $\eta=0.01$ which were used in \cite{biggers2014configuring,rosen2016mobile}).

\textbf{Decorrelated models}.
It has been shown that LDA tends to produce correlated topics when $T$ is too high or too low \cite{cao2009density}.
Therefore, it is interesting to explore models that explicitly attempt to reduce pairwise topic correlations.
The simplest example is TWC-LDA \cite{tan2010topic} which is already implemented in BigARTM library \cite{plavin}.
We consider three possible \texttt{tau} coefficients for decorrelation: $0.02$, $0.05$, $0.1$, while holding \texttt{gamma} equal to $0$.

\textbf{Sparse models}.
Another property of LDA is the difficulty producing sparse models due to the smoothing priors \cite{soleimani14parsimonious,than2012fully}. Some information-based quality metrics treat sparse and smooth topic models differently \cite{than2012fully}, therefore it is important to include sparse models in our analysis.
The simplest sparse model divides its topics into two categories: 
background (general, stopword, un\-in\-for\-ma\-tive, ``slab'') topics and specific (domain, foreground, focused, ``spike'') topics that are sparse compared to the background ones \cite{lin2014dual,fan2019assessing,paul2014discovering,chemudugunta2007modeling}.
The BigARTM supports such models \cite{potapenko2013robust}.
We explore two dataset-adjusted values for smoothing prior and two values for sparsing prior.

In addition, we included a thetaless topic model~(TARTM) designed in~\cite{irkhin2020additive} as an example of a model where sparsity is an emergent property.

\textbf{Sparse decorrelated models}.
For the sake of completeness, we combined different restrictions to obtain a model that is sparse and decorrelated simultaneously.
Additive regularization of topic models (ARTM) allows us to combine several requirements in that manner using a set of regularizers.

\subsection{Corpora Used}

\begin{table}[h]
    \centering
    \caption{Ground Truth on Number of Topics}  
    \begin{tabular}{lllllll}
    \toprule
    Dataset                 & D & W & Expected T & min T & max T \\
    \midrule
    WikiRef220 & $220$ & $4839$ & $5$ & $2$ & $20$   \\
    20NG  & $18846$  & $2174$ & $\numrange{15}{20}$ & $3$ & $40$  \\
    Reuters & $10788$ & $5074$  & $90$ & $5$ & $150$ \\
    Brown  & $500$ & $7409$ & $\numrange{10}{20}$ & $5$ & $25$  \\
    StackOverflow & $895621$ & $3430$ & ?? & $5$ & $60$   \\
    \midrule
    PostNauka & $3404$ & $8417$ & $\numrange{15}{30}$ & $5$ & $50$  \\
    ruwiki-good & $8603$ & $236018$ & $10/90$ & $5$ & $100$  \\
    \bottomrule
    \end{tabular}
\end{table}

We utilize the following datasets: 

\textbf{WikiRef220} \footnote{\url{https://www.multisensorproject.eu/achievements/datasets}} dataset, which firstly appeared in \cite{gialampoukidis2016hybrid}, consists of 220 news articles hyperlinked to a specific Wikipedia article.
The documents are divided into 16 different groups depending on the article linked, but only 5 groups contain more than 5 entries.
Following this line of reasoning, authors describe this dataset as having 5 topics and noise.

\textbf{PostNauka} is a corpus consisting of articles published in ``PostNauka'', a popular Russian online magazine about science\footnote{\url{https://postnauka.ru}}.
The corpus contains 3404 documents.
Each document is labelled with a number of tags, which make it possible to estimate the reasonable number of topics as laying in the range $[10, 30]$.
The previous research on this dataset resulted in a topic model consisting of $19$ topics \cite{intracoh}.

\textbf{20NewsGroups}, \textbf{Brown Corpus} and \textbf{Reuters Corpus} are well-known datasets in NLP.
The general consensus is that 20NG consists of 15-20 topics, Brown consists of 10-20 topics and Reuters consists of 50-100 topics.

\textbf{StackOverflow} is a well-known question and answer site that focuses on programming.
There have been various studies done to find good topics on Stackoverflow for SE \cite{barua2014developers,linares2013exploratory,allamanis2013and,rosen2016mobile}.
We use the already preprocessed version of this corpus from \cite{agrawal2018wrong}, which consists of $\num{895621}$ documents.

\textbf{Russian Wikipedia}.
We introduce a new dataset, which we call ``ruwiki-good'' it can be downloaded through the TopicNet library \cite{bulatov2020topicnet}.
To obtain it, we downloaded a Russian Wikipedia database dump and extracted $8603$ articles falling into either of ``featured'' (\foreignlanguage{russian}{избранные}), ``good'' (\foreignlanguage{russian}{хорошие}) or ``solid'' (\foreignlanguage{russian}{добротные}) assessment grades.
An advantage of this corpus is the existence of a curated hierarchy of labels: each article falls into one of 11 main categories\footnote{
Biology, Geography, Science, Arts, History, Culture and Society, Personalities, Religion and Philosophy, Sports and Entertainment, Technology, Economics}, with each of them being further subdivided into a various number of subcategories (e.g. ``History'' $\rightarrow$ ``History of UK'' $\rightarrow$ ``Murders in the United Kingdom'').
We believe it to be a valuable testing ground for issues regarding topic granularity.

\section{Results and Discussion}
\label{sec:results-and-discussion}

We organized the results of the conducted experiments in Table \ref{tab:metric_performance}.
In order to provide meaningful insight into the performance of the considered metrics, we designed three features to characterize their behaviour.
We wanted to assess metric ability to provide topic number estimation independent from model random initialization, the ``readability'' of obtained plots and precision of the metric providing an expected number of topics.

\begin{table}[ht]
  \centering
  \caption{Metric comparison by applicability averaged over datasets.}
  \begin{tabular}{lccc}
    \toprule
    Score & Jaccard & Informativity & Expected \\
    \midrule
    AIC                  &        0.280 &          \textcolor{blue}{0.542} &     \textcolor{blue}{0.578} \\
    AIC sparse          &        \textcolor{blue}{0.219} &          0.111 &     0.100 \\
    BIC                  &        \textcolor{blue}{0.128} &          \textcolor{blue}{0.444} &     \textcolor{blue}{0.461} \\
    BIC sparse          &        0.274 &          0.164 &     0.128 \\
    MDL                  &        \textcolor{blue}{0.096} &          \textcolor{blue}{0.488} &     \textcolor{blue}{0.414} \\
    MDL sparse           &        0.282 &          \textcolor{blue}{0.428} &     \textcolor{blue}{0.256} \\
        \midrule
    renyi-0.5            &        0.470 &          \textcolor{blue}{0.507} &     \textcolor{blue}{0.425} \\
    renyi-1              &        0.356 &          \textcolor{blue}{0.475} &     \textcolor{blue}{0.394} \\
    renyi-2              &        \textcolor{blue}{0.230} &          0.299 &     0.183 \\
        \midrule
    D-Spectral           &        0.456 &          0.144 &     0.083 \\
    D-avg-COS            &        0.430 &          0.113 &     0.089 \\
    D-cls-COS            &        0.526 &          0.148 &     0.172 \\
    D-avg-L2             &        0.682 &          0.250 &     0.119 \\
    D-cls-L2             &        0.584 &          0.243 &     0.092 \\
    D-avg-H              &        0.356 &          0.062 &     0.089 \\
    D-cls-H              &        0.595 &          0.245 &     0.189 \\
    D-avg-JH             &        0.302 &          0.053 &     0.022 \\
    D-cls-JH             &        0.504 &          0.194 &     0.081 \\
        \midrule
    lift                 &        0.383 &          0.123 &     0.033 \\
    holdout-perplexity   &        \textcolor{blue}{0.228} &          0.025 &     0.019 \\
    perplexity           &        \textcolor{blue}{0.218} &          0.023 &     0.014 \\
    CHI                  &        0.277 &          0.157 &     0.008 \\
    SilhC                &        \textcolor{blue}{0.233} &          0.079 &     0.028 \\
    average coherence    &        0.780 &          \textcolor{blue}{0.472} &     0.208 \\
    uni-theta-divergence &        0.470 &          0.197 &     0.047 \\
    \bottomrule
  \end{tabular}
  \label{tab:metric_performance}
\end{table}

The first column is Jaccard metric calculated the following way: for each random initialization, we extract the optimal value or range of values according to metric specifics.
Then we calculate the Jaccard distance between intersection and union of those sets excluding cases when metric points at the boundaries of the experiment interval.

The second column gives a proportion of how many times the metric results were ``readable'' meaning that they fall in one of the categories:
\begin{itemize}
\item Have a pronounced min/max value/values
\item Have an interval/s around min/max value
\item Have a region of alternating peaks
\end{itemize}
All other types of encountered metric behaviour can be described as either independent from the number of topics or not having any of the described above behaviour (having optimal value outside the range of the experiment).

The last metric is an average of a boolean value: was an expected number of topics in the range of optimal values provided by the metric for this model.
The results in the table \ref{tab:metric_performance} cast doubt on the notion that the number of topics is a well-defined property of a particular corpus (or, at least, that current methods are suitable for deducing it).

\subsection{Common Issues Encountered}

\begin{figure*}[h]
  \centering
  \begin{tabular}{@{}cc@{}}
    \includegraphics[width=\linewidth]{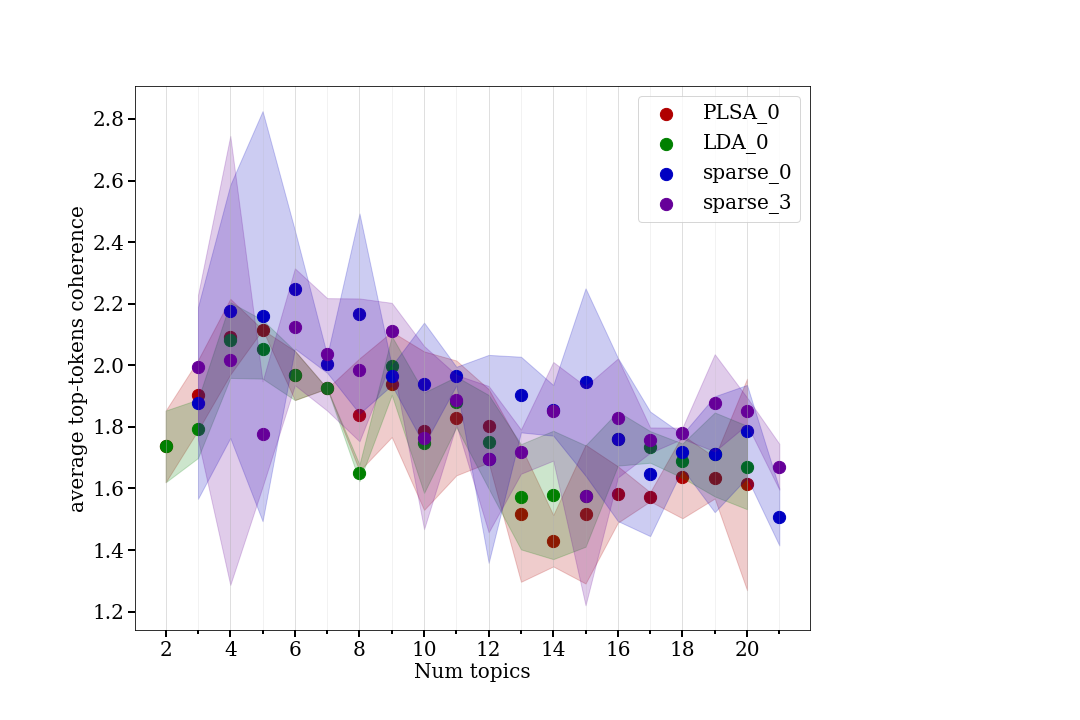} 
  \end{tabular}
  \caption{The average coherence of each topics, $1 < T < 21$. The models depicted are LDA with symmetric prior, LDA with heuristic prior and sparse model 0.}
  \label{fig:wref_coherence1}
\end{figure*}

\begin{figure*}[h]
  \centering
  \begin{tabular}{@{}cc@{}}
    \includegraphics[width=\linewidth]{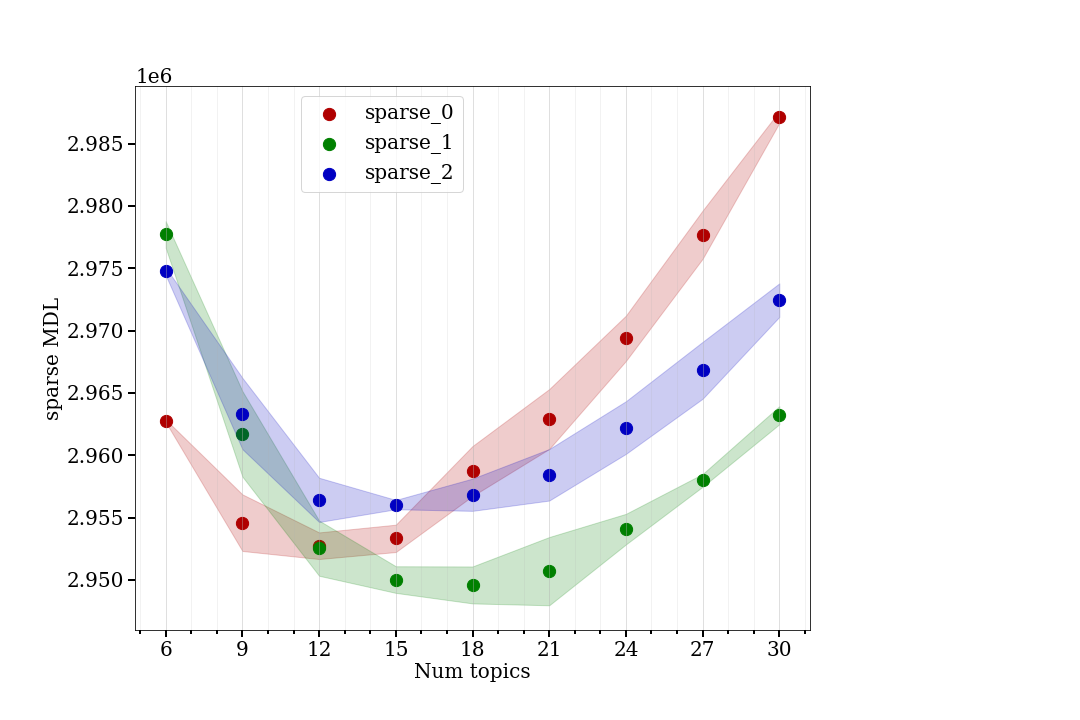} 
  \end{tabular}
  \caption{Sparse MDL criterion for sparse models with different spasity hyperparameter values.}
  \label{fig:wref_coherence2}
\end{figure*}

\textbf{Model Dependency}.
Our first observation is that the ``optimal number of topics'' depends on the model used.
It could be influenced by a particular choice of the topic modelling scheme or even by the hyperparameter configuration within the model scheme.
Figures~\ref{fig:wref_coherence1} and~\ref{fig:wref_coherence2} demonstrate this with WikiRef220, but this is common with other corpora as well.

\textbf{Randomization}.
The second thing to note is the variance caused by random seed (the issue is further complicated if the model dependent on the document order).

As was mentioned earlier, we ran experiments using three random initializations.
If one to look at each curve separately (instead of averaging all three together), it is often the case that their behaviours do not match.
The most frequent case is having partially overlapping and adjacent peaks (e.g. \texttt{seed\_0} gives maximum at 15 topics, \texttt{seed\_1} gives 15 topics as well, but \texttt{seed\_2} gives 14 topics).
The common case of different but adjacent peaks is also easy to analyze, but problematic cases of peaks being significantly separated or only some trajectories having noticeable peaks occur as well.

The natural approach is to build more models with $T$ located in the region of interest, and select the value of $T$ best at average or best in the worst case.
However, this is somewhat self-defeating, since practitioners are usually looking for a single ``best'' topic model.
The approach of building a statistically significant number of different topic models, determining a subset of models with the ``best'' $T$ hyperparameter and then choosing an arbitrary element of that subset, appears flawed in that regard.

The direction indicated by the studies \cite{mehta_clustering_bank,mantyla2018measuring} seems to be a more promising approach for tackling this problem.
First, one needs to create a number of different topic models.
Second, one needs to extract from these topic models a set of topics that are considered ``good'' (coherent, interpretable) or ``strong'' (robust, reproducible).
These topics are saved for the later analysis; some care should be taken to ensure that all saved topics are unique enough.
When this process converges, the set of different topics found that way produces the topic model we are looking for.
The value of $T$ is a by-product of the process.

\textbf{Methods' disagreement}.
The question of agreement could be divided into two:
1) Do different criteria agree with each other?
2) Are variations of the same method consistent? 

\begin{figure*}[h!]
  \centering
  \begin{tabular}{@{}cc@{}}
    \includegraphics[width=\linewidth]{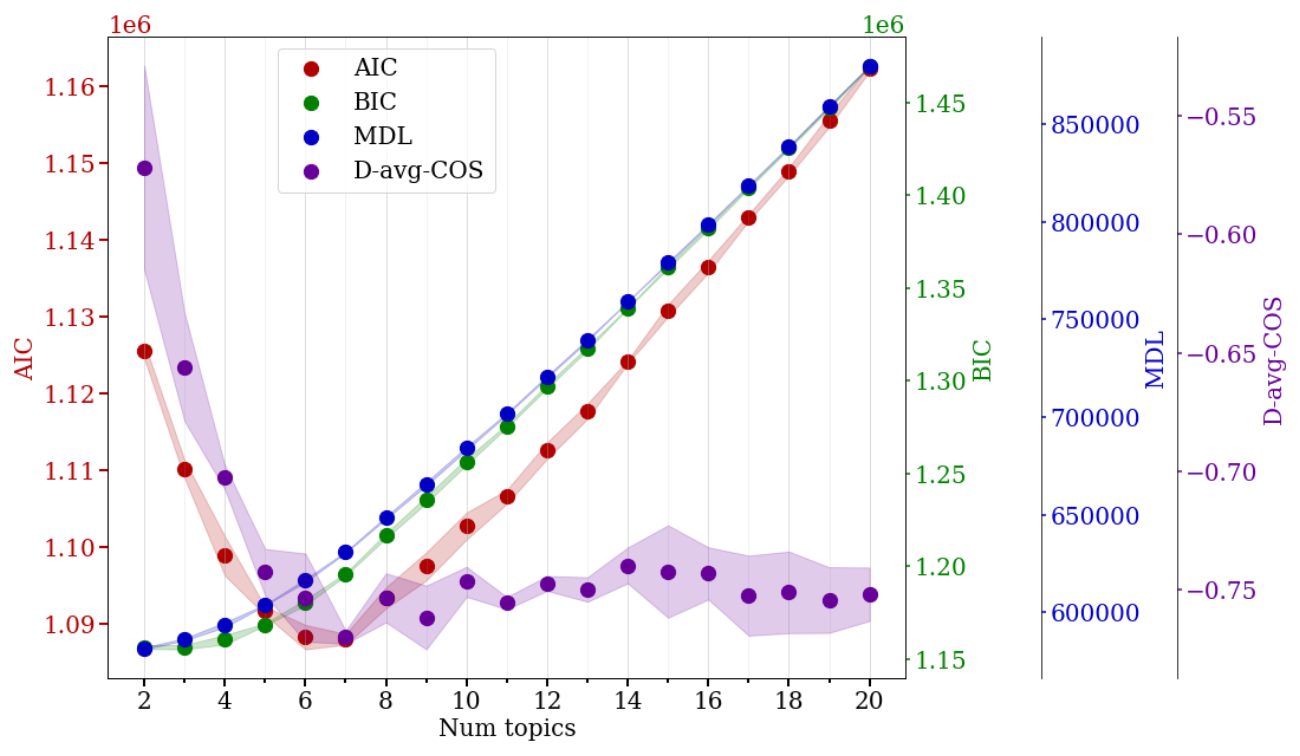} \\
    \includegraphics[width=\linewidth]{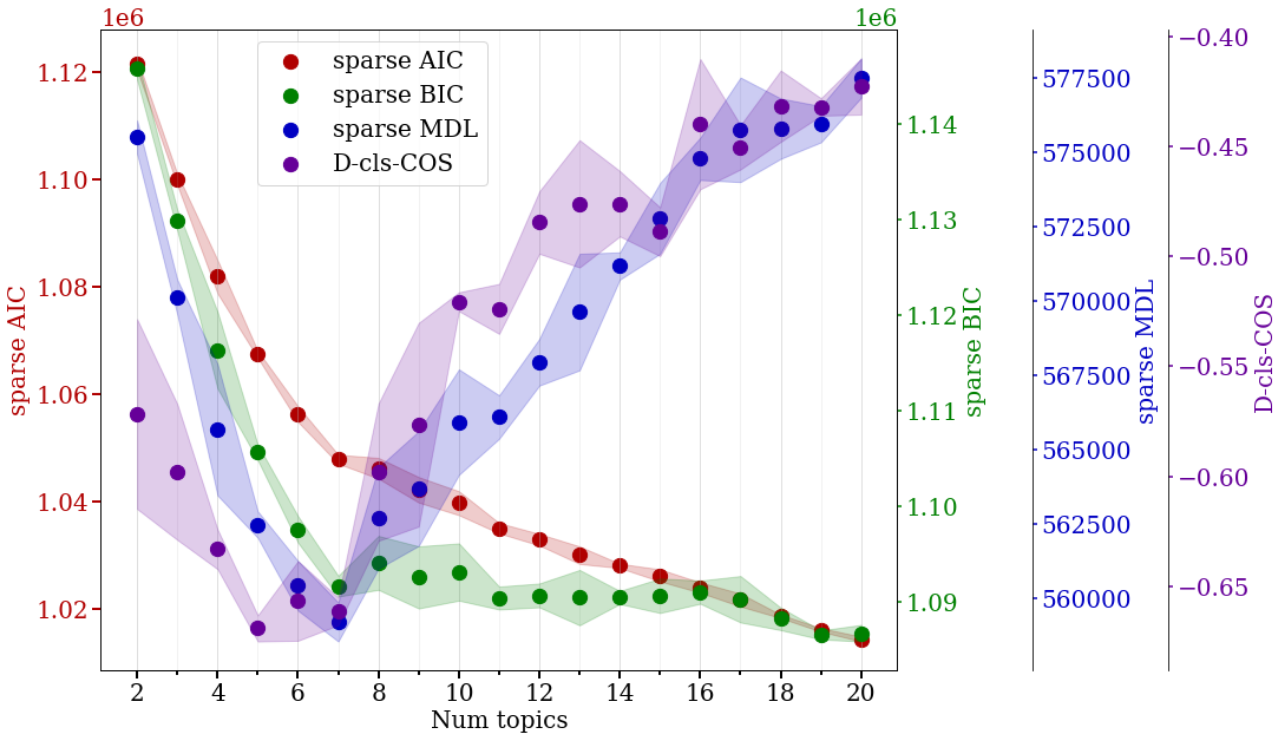} 
  \end{tabular}
  \caption{A set of quality metrics exploring various $T$ for PLSA, $1 < T < 21$. The metrics depicted are AIC, MDL that accounts for model sparsity, and cosine-based diversity (taken with a negative sign, so the minimum corresponds to the ``best'' value).
  We see all metrics agreeing with 7 being a reasonable value for $T$.}
  \label{fig:wref_diversity_overlap}
\end{figure*}

The answer to the first question is negative. Generally, the value determined by diversity-based methods is several times larger than the value determined by other methods.
The differences between other methods are less drastic, but they are frequently significant.
The sole exception is WikiRef220 dataset, where regions overlap, as seen on \ref{fig:wref_diversity_overlap}.

The answer to the second question is inconclusive. The values given by similar methods tend to differ only slightly.
However, it is very common that some criterion gives an answer but a number of variations of it fail.
Therefore, we recommend examining several related measures.

The change of metric in diversity-based methods does not affect the location of peaks, but influences how pronounced they are (which could change the result since we choose the highest peak as the recommended value of $T$).
However this appears to be a result of the same random initialization, as opposed to the existence of the structure inherent in the dataset.
To sum up, the diversity-based methods are robust to the metric change.
That being said, the Euclidean metric appears to be the least informative of all.

\textbf{Objectivity concerns}.
Our approach of checking for sole pronounced minimum/maximum on the plot misses less reliable features of data.
The features such as the location of the first peak, the location where the curve have flattened out, elbow points, and inflection points might be useful in practice.
Unfortunately, using these features introduces too much subjectivity and noise; therefore it is not suitable for selecting the optimal $T$ as an objective, absolute property of the corpus.
We decided to stick to simpler approach for analyzing the $T$ dependency curves.

It might be possible to improve some poor-performing methods by giving explicit guidelines on such cases.
In cases where the curve flattens out, just subtracting some function linear in $T$ is enough to produce a pronounced extremum.
However, this issue is outside the scope of the current study.






\subsection{Properties of the Studied Metrics}

\textbf{Diversity}.
The value of $T$ provided by this method appears to significantly overestimate the number of topics needed.
For the most datasets, the optimum appears to be located outside of  the observed range of numbers of topics.
The curve tends to flatten out instead of showing maximum (which is sometimes remedied by considering the average distance to closest instead of average pairwise distance between all topics).
The location of optimum could change across different random initializations.

\textbf{Information-theoretic}.
These methods are better employed in conjunction, since a single method often fails to produce an estimate of $T$ for some models and datasets.
Taken together, however, they usually give reasonable values of $T$ (which differ from the ``golden standard'', however).
The location of optimum appears remarkably stable across different random initializations. 

\textbf{Entropic}.
These metrics are most likely to give pronounced optima that are robust to random initializations.
However, the location of the minimum significantly depends on the value of $\varepsilon_0$, and the ``default'' value of $\varepsilon_0 = (W)^{-1}$ fails to give expected results.

\textbf{Clustering}.
Silhoutte Coefficient and Calinski--Harabasz index almost always fail to provide any estimate of $T$.
We suspect that the feature space induced by $\Theta$ is not particularly suitable for cluster analysis.
This is further supported by work of \cite{krasnov19clustering}.

\textbf{Spectral Divergence}.
This method is very noisy and difficult to employ.
It is unable to provide any estimate when applied to sparse models (the curve monotonously decreases).

\textbf{Coherence}.
This method is very noisy and difficult to employ.
We observe that average coherence is a declining function of $T$ with frequent fluctuations.
Hence, it is common to find global maximum reached on the small $T$, making this method ill-suitable for $T$ selection.

\textbf{Perplexity}.
The perplexity appears to be monotonous in almost every case, without any notable features helpful for selecting $T$.
This behaviour contradicts earlier works where held-out perplexity had pronounced local minimum.
We conjecture that it depends on precise implementation details, such as treatment of out-of-vocabulary words.

\begin{figure*}[h]
  \centering
  \begin{tabular}{@{}cc@{}}
    \includegraphics[width=\linewidth]{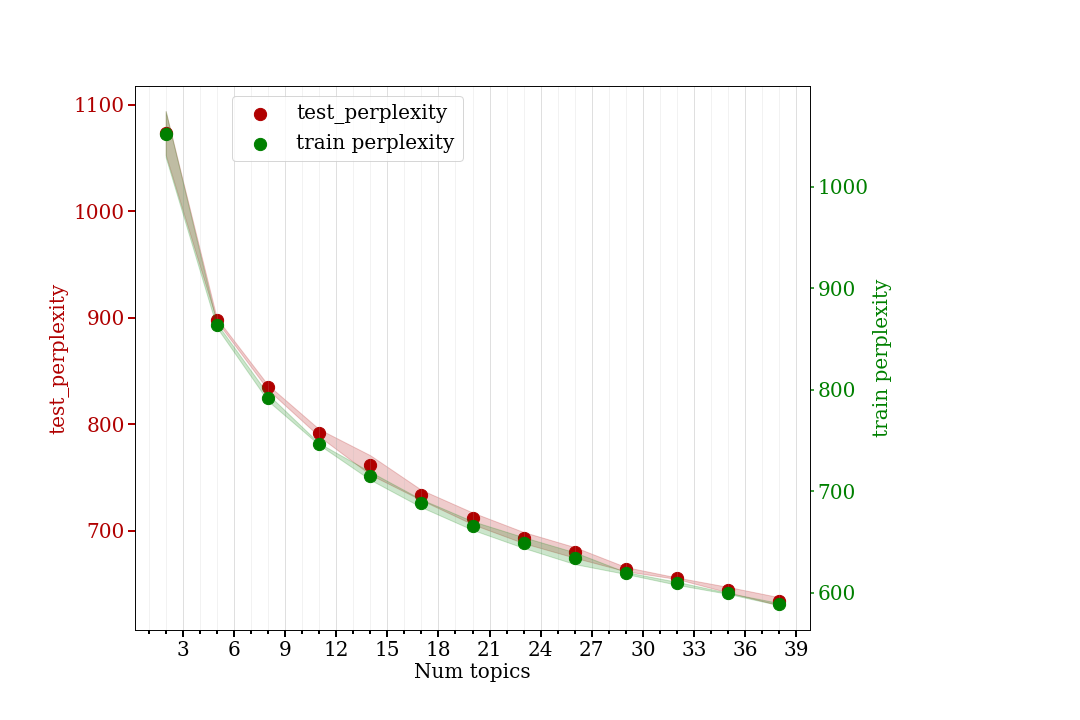} 
  \end{tabular}
  \caption{Comparision of holdout perplexity and train perplexity for LDA model. Similar behaviour was observed for all considered datasets.}
  \label{fig:wref_perp}
\end{figure*}

Rate of perplexity change (RPC) often produces stable peaks and plateaus, but they do not coincide with the expected value.
The numeric result is highly model-dependent.

\begin{figure*}[h]
  \centering
  \begin{tabular}{@{}cc@{}}
    \includegraphics[width=\linewidth]{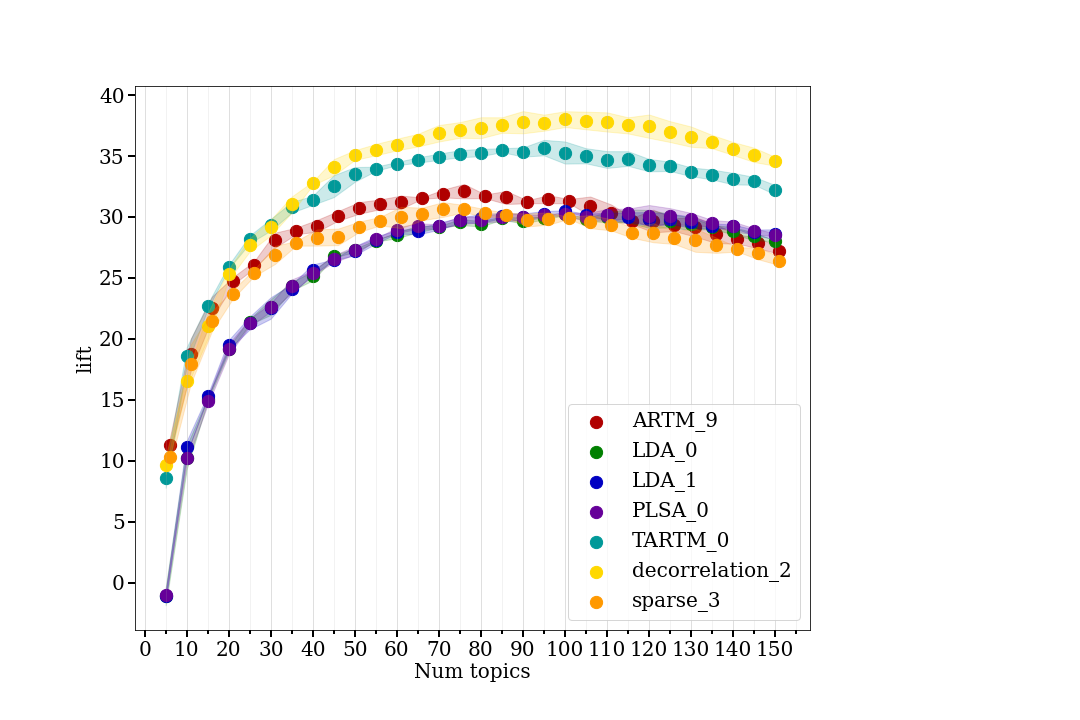} 
  \end{tabular}
  \caption{Multiple maxima of the Lift metric for different types of topic models.}
  \label{fig:wref_lift}
\end{figure*}

\textbf{Lift}.
This metric indicates the optimal topic number as a maximum value from the plot.
Across most datasets, the maximal value reached on the biggest topic model in the experiment well outside the expected optimal number for the datasets.
However, for the Stack Overflow datasets, we observed pronounced maxima for different model families \ref{fig:wref_lift}.

We suspect this was achieved by aggressive token filtration of the original dataset.

\textbf{Stability}.
Authors of this approach suggest looking for a minimal value or one of the multiple local minima as an indicator for optimal number of topics.
However in our experiments we did not always observed the desired behaviour on all datasets and had to settle for a plateau, or just a decline in the rate of instability increase.
We also observed the following problems with the metric.
First, it becomes too noisy for the models with small number of topics
(less than $\numrange{10}{15}$).
Secondly, this metric is not well-suited to deal sparse models: the obtained results are too noisy to conclude the number of topics. 
Thirdly, an estimate of the optimal number of topics given by this method usually was lower than the expected number of topics.


\section{Conclusion}

Further analysis demonstrates that intrinsic methods are far from being reliable and accurate tools in the search of optimal number of topics. 

From our experiments, we see that the best performance was achieved by the simplest approaches: AIC, BIC, MDL, renyi.
Those metrics provide their judgement based on rough estimation of the model state unlike their counterparts deriving their value from finer topic-level structure of the models.
The more intricate methods (lift, coherence, diversity) that attempt to directly measure some desirable qualities of topic model fail to give satisfactory results.

We see that many approaches provide a set of solutions or even a range of optimal number of topics.
This contradicts the naive notion of optimal $T$ being a single fixed value attached to a corpus.
This behaviour can point at the other problem in the field: a concept of topic is ill-defined from granularity point.
Every topic could be divided into subtopics, possibly, without worsening intrinsic metric such as topics' distinctness or clustering quality. Noting that we failed to find methods agreeing with each other on almost all datasets we conclude that the notion of ``optimal number of topics'' may have a few myths in it requiring a deeper consideration.

A text-book approach for the topic model to learn ``latent'' word-topic distributions leads to perception that topics exist within a collection and can be found no matter the approach for finding them.
According to this point of view a dataset is defining the true number of distributions generating it.
This number should not change from type of model extracting those distributions or a type of intrinsic metric finding this number to be ``optimal''.
However, in our experiments, we see that this is not true.
If anything the number of topics is mainly a model-dependent quantity, and partially defined by the implemented optimal topic number approach.
If there is a lesson to be learned from the data that topic number has been misinterpreted by the community and its just another machine learning model hyperparameter to be tuned.

In light of this consideration we see a few ways how the community already deals with that problem:
\begin{itemize}
\item Selecting a model according to secondary task.
\item Building a hierarchy of topics and pruning it afterwards.
\item Improving the process of human (semi-) supervision. 
An example of this approach would be the suggestion from \cite{tang14look}: employing a weak supervision from users to fine-tune threshold hyperparameter that determines the number of topics.
\end{itemize}

We also suggest looking in the following directions as they might turn out to be fruitful in dealing with the question:
\begin{itemize}
\item Eliminating the $T$ hyperparameter. 
\item Developing topic models more robust to the change of $T$.
For example, consider a hypothetical procedure allowing one to build a topic model for a given number of topics in which all topics would be interpretable.
Such procedure would render the question of $T$ determination largely irrelevant.
\item New event detection algorithm with subsequent automatic change of $T$.
\end{itemize}

The ways we considered in this paper to evaluate this ``optimal number of topics'' using intrinsic model quality criteria do not inspire confidence.
We conclude that the optimal number of topics depends not so much on the dataset as on the very method of determining the number of topics itself and the topic model being used, or even on the purpose for which topic modeling is applied.


While previously proposed methods remain work of fiction rather than real algorithms we suggest our readers to treat $T$ merely as another model hyperparameter.
Our findings suggest that practitioners should not try to estimate ``natural'' number of topics inherent in corpus; instead, one should focus on the questions similar to the following:

\begin{itemize}
\item How many documents should a topic consist of, on average?
\item Which degree of granularity is desired?
\item If external labels are available: how to incorporate this additional knowledge into the model?
\item Are inferred topics unique enough?
\end{itemize}

In the end, we conclude that the main purpose of topic modeling should not be the search for the optimal number of topics, but the search for such a method of model training which, given the number of topics, results in a model whose topics in the absence of external criterion are all interpretable.



\bibliographystyle{splncs04}
\bibliography{biblio}


\end{document}